\documentclass[journal,twoside,web]{ieeecolor}
\usepackage{generic}
\usepackage{cite}
\usepackage{amsmath,amssymb,amsfonts}
\usepackage{algorithmic}
\usepackage{graphicx}
\usepackage{algorithm,algorithmic}
\usepackage{hyperref}
\hypersetup{hidelinks}
\usepackage{textcomp}
\usepackage{CJKutf8}
\usepackage{booktabs}
\usepackage{multirow}
\usepackage{multicol}
\usepackage{longtable}
\usepackage{makecell}
\usepackage[misc]{ifsym}
\usepackage{xcolor, colortbl}

\def\BibTeX{{\rm B\kern-.05em{\sc i\kern-.025em b}\kern-.08em
    T\kern-.1667em\lower.7ex\hbox{E}\kern-.125emX}}
\begin{document}
\begin{CJK}{UTF8}{gbsn}
\definecolor{mygray}{gray}{.9}
\definecolor{mypink}{rgb}{.99,.91,.95}
\definecolor{mygreen}{rgb}{.9,.99,.9}

\title{BianCang: A Traditional Chinese Medicine Large Language Model}

\author{Sibo Wei\textsuperscript{†}, Xueping Peng\textsuperscript{†}, \IEEEmembership{Senior Member, IEEE}, Yifei Wang, Tao Shen, Jiasheng Si, Weiyu Zhang \IEEEmembership{Member, IEEE}, Fa Zhu, Athanasios V. Vasilakos, \IEEEmembership{Senior Member, IEEE}, Wenpeng Lu\textsuperscript{*}, \IEEEmembership{Member, IEEE}, Xiaoming Wu and Yinglong Wang\textsuperscript{*}
\thanks{Sibo Wei, Jiasheng Si, Weiyu Zhang, Wenpeng Lu, Xiaoming Wu and Yinglong Wang are with the Key Laboratory of Computing Power Network and Information Security, Ministry of Education,
Shandong Computer Science Center (National Supercomputer Center in Jinan),
Qilu University of Technology (Shandong Academy of Sciences), 
and with Shandong Provincial Key Laboratory of Computing Power Internet and Service Computing, Shandong Fundamental Research Center for Computer Science, 
Jinan, China (e-mail: sibo.wei@foxmail.com, \{jiashengsi, zwy, wenpeng.lu, wangyinglong\}@qlu.edu.cn, wuxm@sdas.org).}
\thanks{Xueping Peng and Tao Shen are with the Australian Artificial Intelligence Institute, University of Technology Sydney, Sydney (e-mail: \{xueping.peng, tao.shen\}@uts.edu.au).}
\thanks{Yifei Wang is with the Affiliated Hospital of Shandong University of Traditional Chinese Medicine, Jinan, China (e-mail: 71000686@sdutcm.edu.cn).}
\thanks{Fa Zhu is is with the College of Information Science and Technology \& Artificial Intelligence, Nanjing Forestry University, Nanjing, China (E-mail: fazhu@njfu.edu.cn).}
\thanks{Athanasios V. Vasilakos is with the Department of ICT and Center for AI Research, University of Agder (UiA), Grimstad, Norway (E-mail: thanos.vasilakos@uia.no).}
\thanks{\textsuperscript{†}Sibo Wei and Xueping Peng contributed equally to this work. }
\thanks{\textsuperscript{*}Wenpeng Lu and Yinglong Wang are the corresponding authors.}
}

\maketitle

\begin{abstract}
The surge of large language models (LLMs) has driven significant progress in medical applications, including traditional Chinese medicine (TCM). However, current medical LLMs struggle with TCM diagnosis and syndrome differentiation due to substantial differences between TCM and modern medical theory, and the scarcity of specialized, high-quality corpora. 
To this end, in this paper we propose BianCang (扁仓)\footnote{
The term \textit{BianCang} (扁仓) is formed by combining the first characters of the names of two renowned ancient Chinese TCM physicians, \textit{Bian Que} (扁鹊) and \textit{Cang Gong} (仓公).
}, a TCM-specific LLM, using a two-stage training process that first injects domain-specific knowledge and then aligns it through targeted stimulation to enhance diagnostic and differentiation capabilities. Specifically, we constructed pre-training corpora, instruction-aligned datasets based on real hospital records, and the ChP-TCM dataset derived from the Pharmacopoeia of the People's Republic of China. We compiled extensive TCM and medical corpora for continual pre-training and supervised fine-tuning, building a comprehensive dataset to refine the model's understanding of TCM. Evaluations across 11 test sets involving 31 models and 4 tasks demonstrate the effectiveness of BianCang, offering valuable insights for future research. Code, datasets, and models are available on GitHub.\footnote{\href{https://github.com/QLU-NLP/BianCang}{https://github.com/QLU-NLP/BianCang}}
\end{abstract}


\begin{figure}[t]
\centering
\includegraphics[width=0.42\textwidth]{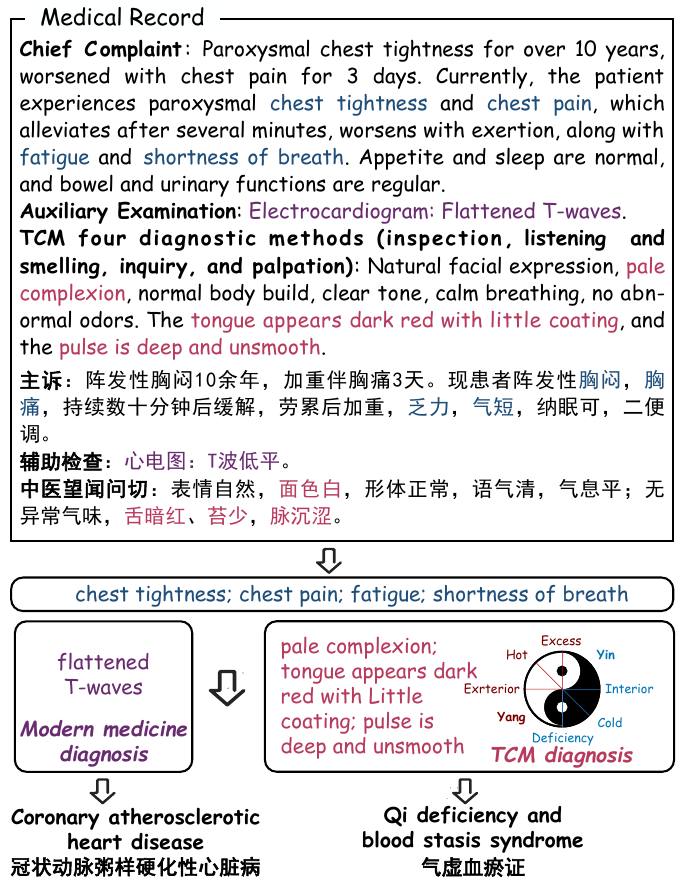}
\caption{\small Different diagnostic processes of TCM and modern medicine for the same medical record. Modern medicine relies on the chief complaint and auxiliary examination on the T-wave through electrocardiograms, diagnosing the patient based on specific value changes and trends. In contrast, TCM interprets the patient's chief complaint and diagnostic information from the four diagnostic methods within a unique Yin-Yang framework, identifying underlying causes and synthesizing the findings to determine the syndrome type. While modern medicine depends on quantifiable data, TCM is more abstract and experience-based.}
\label{fig:case}
\end{figure}

\section{Introduction}

Recently, LLMs have achieved remarkable success in natural language processing \cite{zhao2023survey, li2024fundamental}. General-purpose models such as DeepSeek-V3~\cite{liu2024deepseek}, DeepSeek-R1~\cite{guo2025deepseek}, GPT-4 \cite{achiam2023gpt} and Qwen2.5~\cite{yang2023baichuan} have demonstrated fascinating capabilities across diverse tasks. 
Building on these models, researchers have effectively applied LLMs to vertical domains such as healthcare \cite{liu2025generalist,luo2024biomedgpt,zhao2024automated}, law \cite{fei-etal-2024-lawbench}, and finance \cite{yang2024finrobot}, often at relatively low cost by prompting, in-context learning, or continual training. Among verticalizations, healthcare stands out due to its critical role in improving human health and enhancing longevity, attracting increasing interest from both academia and industry \cite{lan2025large, wu2024callm}.

The major challenge to adapting general LLMs into the biomedical domain is a lack of accurate knowledge and expertise \cite{he2023survey}, which is critical to tackle a variety of domain-specific tasks. 
In general, works addressed this challenge by leveraging continual/secondary training methods through reinforcement learning and streamlined fine-tuning \cite{wang2023huatuo, chen2023huatuogpt} and improving data quality with specialized datasets \cite{bao2023disc, luo2024taiyi}.

Despite successful applications of the resulting medical LLMs in modern medicine, the performance of TCM still lags far behind, mainly due to theoretical differences between TCM and modern medicine, leading to further knowledge and domain gaps. To be specific, TCM has a unique and complete theoretical foundation. Unlike modern medicine, which treats diseases based on their types, TCM determines the patient's syndrome type by analyzing the evidence collected through the ``four diagnostic methods'' (望/闻/问/切: inspection, listening and smelling, inquiry, and palpation). 
Treatment is then administered based on the syndrome type. As a result, patients with the same disease and different syndrome types may receive different treatments, while patients with different diseases and the same syndrome type may receive the same treatment. This approach, known as ``treating the same disease with different therapies'' (同病异治) and ``treating different diseases with the same therapy'' (异病同治), is central to TCM \cite{mucheng2022tcm}. 

Take Fig.~\ref{fig:case} as an example: For the same patient, modern medicine uses medical instruments to measure the T-wave (related to changes in the electrocardiogram) and diagnoses the patient based on the specific value changes and trends of the T-wave. In contrast, TCM maps the patient's described symptoms and the evidence gathered from the four diagnostic methods into a unique Yin-Yang (阴-阳) coordinate system, analyzes the underlying causes, and synthesizes these findings to determine the patient's syndrome type. 
As such, TCM applications are more abstract, experience-based, and hard to explain \cite{mucheng2022tcm}, differing from modern medicine relying on measurable data from instruments. 

Recently, works have explored the development of TCM-specific LLMs, including Qibo \cite{jia2025qibo}, which was trained on a specialized TCM corpus for TCM consultations, Zhongjing \cite{yang2024zhongjing}, which focused on doctor-patient dialogues to improve proactive inquiry and consultation understanding, and Lingdan \cite{hua2024lingdan}, which aimed to enhance the model's grasp of TCM and pharmacology.

However, these models still face significant challenges in practical syndrome differentiation and diagnostic analysis, where the complex and abstract nature of TCM theories and the scarcity of high-quality, structured diagnostic data hinder their ability to perform accurate and consistent syndrome differentiation in real-world diagnostic scenarios.

To address the challenge above,
we propose BianCang (扁仓), a large TCM language model built on Qwen-2/2.5, designed to enhance syndrome differentiation and diagnostic capabilities in TCM. BianCang employs a two-stage training process: in the first stage, extensive traditional Chinese medicine and medical knowledge are integrated through continual pre-training of the foundational model, whereas in the second stage, supervised fine-tuning is conducted to activate and align the model's internal knowledge. 
We evaluated BianCang on 11 benchmark datasets across 4 tasks, and the results demonstrate that it consistently outperforms existing open-source TCM-specific and Chinese medical language models across all key performance dimensions.
\begin{figure*}[htbp]
\centering
\includegraphics[width=1\textwidth]{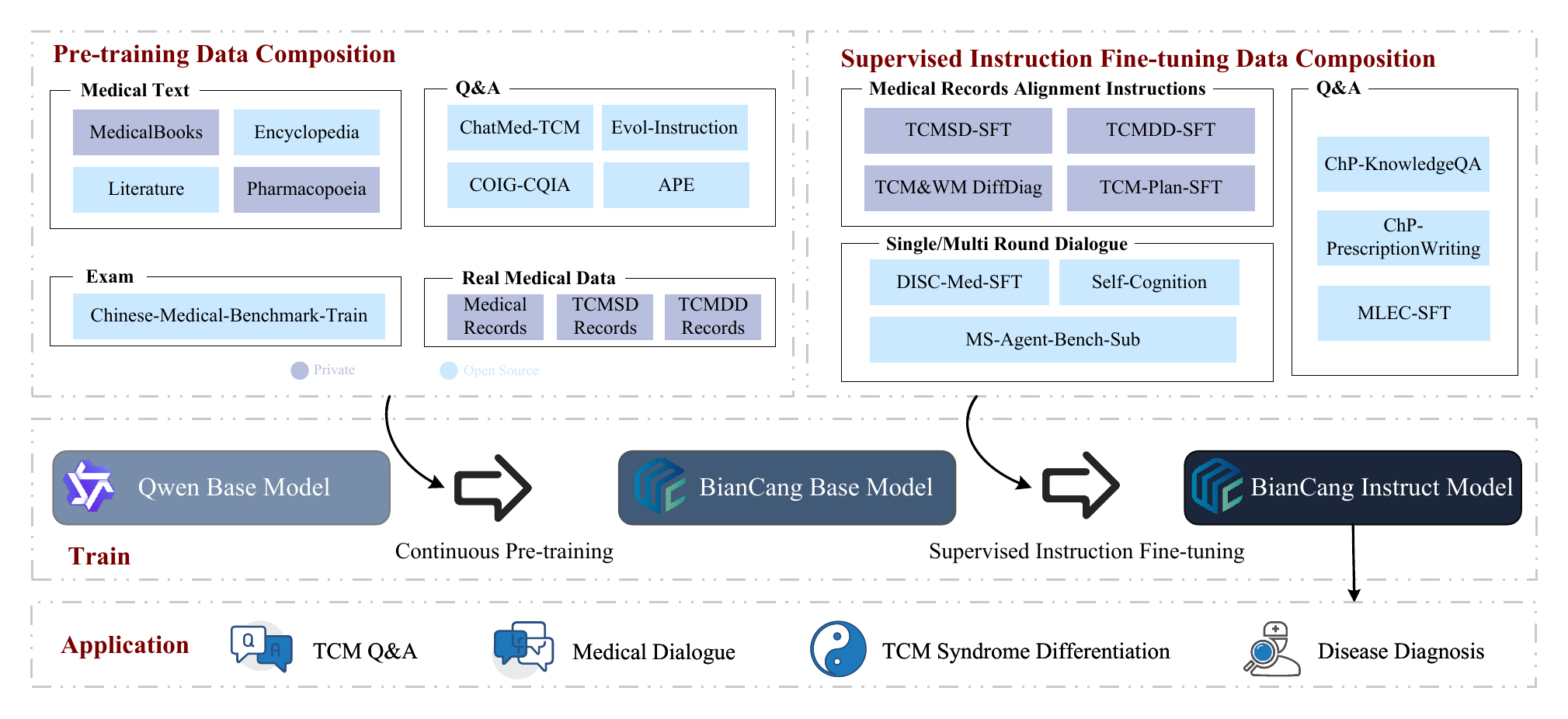}
\caption{The overall flowchart of constructing BianCang. In the first stage, extensive traditional Chinese medicine and medical knowledge is injected into the foundational model through continual pre-training. In the second stage, supervised fine-tuning is applied to activate and align the internal knowledge of the model.}
\label{fig:model}
\end{figure*}
Our main contributions are summarized as follows:
\begin{itemize}
    \item We presented BianCang, a new TCM LLM with a two-stage training process tailored to the distinctive characteristics of TCM. BianCang has acquired comprehensive TCM knowledge and real clinical expertise, demonstrating strong and consistent capabilities in syndrome differentiation and disease diagnosis.
    \item We constructed and open-sourced the ChP-TCM dataset, based on the Pharmacopoeia of the People's Republic of China, specifically designed for pre-training and supervised fine-tuning of TCM models.
    \item We performed a comprehensive evaluation of existing medical and TCM LLMs from multiple perspectives, considering both objective and subjective aspects. Extensive experiments involving 11 datasets, 31 models, and 4 tasks validate the effectiveness and robustness of the BianCang model, providing valuable benchmarks for future research.  
\end{itemize}

\section{Related Work}

\subsection{LLMs in Chinese Medical Domain}
Among the existing Chinese medical LLMs, some works are based on open-source foundation models, which are then fine-tuned on self-constructed medical datasets. Examples include DoctorGLM \cite{xiong2023doctorglm} and BenTsao \cite{wang2023huatuo}. To further enhance the capabilities of medical LLMs, researchers have explored optimizing training methods from various perspectives. For instance, Zhang et al. introduced RLMF (Reinforcement Learning from Mixed Feedback) in their training process to coordinate both ChatGPT-extracted data and real-world data from doctors, resulting in HuatuoGPT1 \cite{zhang2023huatuogpt}. Chen et al., aiming to reduce the complexity of multi-stage training strategies, proposed a one-stage domain adaptation protocol where heterogeneous data from both traditional pre-training and supervised stages are unified into a simple instruction-output pair format, which they used to train HuatuoGPT2 \cite{chen2023huatuogpt}. Tian et al. developed ChiMedGPT, following a process that included pre-training, supervised fine-tuning (SFT), and reinforcement learning from human feedback (RLHF) \cite{chimed2024tian}. Other researchers have attempted to improve the quality of training data. For example, Bao et al. constructed a high-quality SFT dataset using three strategies: utilizing medical knowledge graphs, reconstructing real-world dialogues, and incorporating human-guided preference rephrasing, significantly enhancing the model's response capabilities in both single-turn and multi-turn consultation scenarios \cite{bao2023disc}. Luo et al. built a rich, multi-task bilingual dataset in both Chinese and English and trained the Taiyi model \cite{luo2024taiyi}. Although these medical LLMs have achieved considerable success in modern medicine, their performance in the TCM domain is significantly lacking. 

\subsection{LLMs in TCM Domain}
Previous Chinese medical LLMs have mostly focused on modern medical systems. Unlike modern medicine, TCM emphasizes a holistic view and individualized treatment, focusing on balancing Yin-Yang, harmonizing Qi (气) and blood (血), and coordinating the internal organs. TCM diagnoses are conducted through the four diagnostic methods (inspection, listening/smelling, inquiry, and palpation) to understand the patient's overall condition \cite{mucheng2022tcm}. 
The theoretical foundation of syndrome differentiation and disease diagnosis in TCM is unique and fundamentally different from modern medicine. Existing medical LLMs are unable to grasp the principles of TCM, resulting in unsatisfactory performance.

Some studies have specifically aimed at improving the capabilities of LLMs in TCM. For example, Zhang et al. built a TCM-specific corpus and trained the Qibo model, which was applied to the pipeline of TCM consultations \cite{jia2025qibo}. Yang et al. constructed a real-world, multi-turn Chinese medical dialogue dataset, CMtMedQA, and trained the Zhongjing model using continual pre-training, SFT, and RLHF, which enhanced the model's ability in active inquiry and multi-turn understanding within the context of TCM \cite{yang2024zhongjing}. Hua et al. focused on traditional Chinese medicine, constructing datasets for TCM pre-training, traditional Chinese patent medicine (TCPM) Q\&A, and the spleen and stomach herbal prescription recommendation, and used these datasets to train the Lingdan model \cite{hua2024lingdan}. Although these efforts have made significant progress in TCM LLMs, these models still lack real-world diagnostic experience in TCM and perform poorly on real-world syndrome differentiation and disease diagnosis tasks.

To address the aforementioned challenges, we propose BianCang, a novel TCM LLM based on Qwen2/2.5. BianCang first undergoes continual pre-training to inject large amounts of TCM and medical knowledge, as well as real medical records, into the model. It is then fine-tuned using a carefully curated set of TCM-specific instructions to activate the model's internal knowledge and align these instructions with the model's internal knowledge. This training strategy ensures consistency in the model's knowledge before and after SFT, significantly improving its performance in tasks such as syndrome differentiation and disease diagnosis in TCM.

\section{Methods}
The real hospital records used in this study were approved by the Ethics Committee of the Affiliated Hospital of Shandong University of Traditional Chinese Medicine (No. AF/SC-08/03.0).
The construction of BianCang consists of two stages: continual pre-training and supervised fine-tuning. From the perspective of cognitive science, this two-stage framework—knowledge injection during pre-training and knowledge alignment during fine-tuning—closely mirrors the ACT-R \cite{anderson2004integrated} cognitive architecture. In ACT-R, declarative knowledge (i.e., factual information) is stored in long-term memory in the form of chunks, while procedural knowledge guides task-specific behavior. Pre-training resembles the accumulation of declarative chunks, whereas fine-tuning acts as a process of chunk activation and proceduralization, enabling the model to use its internal knowledge in a goal-directed manner. This theoretical framing helps justify why pre-training and fine-tuning must work together to achieve effective domain adaptation in large language models. The overall process is shown in Fig. \ref{fig:model}. 

\begin{table*}[h]
\caption{The statistics of the continual pre-training data.}
\begin{center}
\resizebox{0.9\textwidth}{!}{
\begin{tabular}{llrrr}
\hline \bf Name & \bf Type & \bf \# of Tokens & \bf \# of Instances & \bf Size \\ \hline
MedicalBooks & Textbook & 76M & 140,909 & 263MB \\
Encyclopedia & Q\&A & 113M & 411,183 & 460MB \\
Literature & Q\&A & 48M & 177,261 & 212MB \\
Pharmacopoeia & Knowledge Base & 3M & 2,209 & 10MB \\
MedicalRecords & Medical Record & 23M & 12,259 & 73MB \\
TCMSD\&DD Records & Medical Record & 24M & 56,928 & 82MB \\
TCMSyndromeKnowledge & Knowledge Base & 380K & 1,174 & 1MB \\
CMB-Train & Exam & 22M & 269,359 & 73MB \\
ChatMed-TCM & Q\&A & 25M & 112,565 & 104MB \\
COIG-CQIA & Q\&A & 31M & 44,694 & 131MB \\
APE-210K & Math & 13M & 210,488 & 37MB \\
Evol-Instruction-66K & Code & 28M & 66,862 & 117MB \\ \hline
\bf Total & - & 460M & 1,505,891 & 1,563MB \\
\hline
\end{tabular}
}
\end{center}
\label{tab:pretrain-data}
\end{table*}

\subsection{Continual Pre-training}
Recent research indicates that supervised fine-tuning (SFT) is not an effective mechanism of learning domain-specific world knowledge, but rather a process of aligning instructions with the knowledge already embedded within large language models \cite{learning2024ren}. Therefore, we should strive to inject extensive knowledge of traditional Chinese medicine into the foundational model during the pre-training phase \cite{zhou2024lima}.

Books serve as a high-quality source of data for training large language models \cite{gunasekar2023textbooks,dong2024abilities}. To construct a comprehensive corpus of Traditional Chinese Medicine (TCM) and general medical knowledge, we collected a substantial number of digitized books. These texts were processed using Optical Character Recognition (OCR), followed by a cleaning pipeline that applied heuristic rules to remove noisy characters and normalize text formatting. Manual sampling was performed to verify the quality of the processed content. For text segmentation, we employed the LlamaIndex\footnote{\url{https://github.com/run-llama/llama_index}} framework with a sliding-window strategy (i.e., overlapping chunking), using a chunk size of 1024 tokens to maintain contextual coherence.

In addition to textbook data, we incorporated open-source medical encyclopedias, scientific literature, and the ChatMed-TCM dataset \cite{chen2023huatuogpt,zhu2023ShenNong}. To further enrich the model’s TCM-specific knowledge, we integrated authoritative resources such as the Pharmacopoeia of the People's Republic of China (Part I: Traditional Chinese Medicine) and commonly used TCM syndrome knowledge bases. These resources were manually inspected and cleaned on a file-by-file basis to ensure accuracy and consistency.

Recognizing that practical clinical experience is indispensable to TCM reasoning, we further included 12,259 real-world patient case records and 56,928 TCM syndrome differentiation and diagnosis records into the pre-training corpus. Lastly, to enhance the model’s performance on domain-specific assessments, we incorporated the training split of the Chinese Medical Benchmark (CMB) \cite{wang2024cmb}, which evaluates model capabilities across various real-world clinical scenarios.

Besides TCM and medical corpora, we also incorporated general domain corpora into the pre-training data to prevent the model from overfocusing on specific domains, which could lead to catastrophic forgetting in general domains. Among them, COIG-CQIA is a high-quality, multi-domain instruction fine-tuning dataset aligned with human interaction behaviors \cite{bai2024coig}. APE-210K is a diverse collection of mathematical word problems \cite{zhao2020ape210k}. Evol-Instruction-66K is a filtered and curated dataset for code generation \cite{luowizardcoder}. We concatenated the instructions and responses from these three datasets as the pre-training data. Notably, all three general-domain datasets exhibit a certain level of logical reasoning, which we speculate could benefit tasks such as TCM syndrome differentiation, as it also requires logical reasoning skills. 

The data statistics for pre-training are shown in Table \ref{tab:pretrain-data}. Using this data, we performed continual pre-training on Qwen2/2.5-7B and Qwen2.5-14B \cite{yang2024qwen2,yang2024qwen25}, eventually obtaining the BianCang TCM base model. 

The negative log-likelihood (NLL) loss for continual pre-training of a language model can be defined as follows:
\begin{align}
    &\mathcal{L}_{\text{CPT}}(\theta) = \\
    &\notag~~~~- \frac{1}{N} \sum_{i=1}^{N} \sum_{t=1}^{T_i} \log P_\theta(x_t^{(i)} \mid x_1^{(i)}, \ldots, x_{t-1}^{(i)}),
\end{align}
where \( N \) denotes the number of training sequences (documents, sentences, etc.), \( T_i \) denotes the length of the \(i\)-th sequence, \( x_t^{(i)} \) denotes the token at position \( t \) in the \( i \)-th sequence, \( P_\theta(x_t^{(i)} \mid x_1^{(i)}, \ldots, x_{t-1}^{(i)}) \) denotes the probability of the token at position \( t \) given the previous tokens, computed by the model with parameters \( \theta \).

\subsection{Supervised Fine-Tuning}
\begin{table*}[h]
\caption{The statistics of the supervised fine-tuning.}
\begin{center}
\resizebox{0.9\textwidth}{!}{
\begin{tabular}{llrrr}
\hline \bf Name & \bf Type & \bf \# of Tokens & \bf \# of Instances & \bf Size \\ \hline
ChP-KnowledgeQA& Knowledge Base Q\&A& 1M& 7,892& 5MB\\
ChP-PrescriptionWriting& Knowledge Base Q\&A& 338K& 2,512& 1MB \\
DISC-Med-SFT& Single/Multi Round Dialogue& 147M& 464,620& 760MB \\
MLEC-SFT& Knowledge Base Q\&A& 12M& 108,988& 39MB \\
MS-Agent-Bench-Sub& General Q\&A& 21M& 65,000& 103MB \\
TCMSD-DD-SFT& Medical Record Instruction& 25M& 55,428& 91MB \\
Self-Cognition& Single Round Dialogue& 7K& 113& 41KB \\
TCM-WM-DiffDiag-SFT& Medical Record Instruction& 375K& 1,951& 1MB\\
TCM-Plan-SFT& Medical Record Instruction& 21M& 13,409& 70MB\\ \hline
\bf Total& - & 228M & 719,913 & 1,070MB \\
\hline
\end{tabular}
}
\end{center}
\label{tab:sft-data}
\end{table*}

SFT is the core stage for applying the large language model of TCM to real-world scenarios, and the key phase in enabling the model's conversational abilities. The quality and diversity of the instruction data are crucial to the SFT process \cite{liu2024deepseek}. To maximize the activation of the knowledge embedded in the pre-trained model and align it with downstream real-world TCM tasks, we constructed a high-quality and diverse instruction dataset, containing the following five types of data:

\textbf{TCM Pharmacopoeia Data}. During the CPT stage, the raw text from the Pharmacopoeia of the People’s Republic of China (Part I: Traditional Chinese Medicine) was directly used to inject domain-specific knowledge into the model. To further enhance the model’s TCM instruction-following ability, we constructed two supervised instruction datasets—ChP-KnowledgeQA and ChP-PrescriptionWriting—based on this source. The ChP-KnowledgeQA dataset includes question-answer pairs covering topics such as the properties, identification, processing methods, taste, meridian tropism, functions, indications, dosage, and storage of TCM herbs. The ChP PrescriptionWriting dataset consists of QA pairs focused on composing and formulating TCM prescriptions. To ensure diversity, we manually designed a set of query templates when constructing these datasets. Together, these datasets constitute the ChP-TCM dataset, used exclusively in the supervised fine-tuning stage.

\textbf{Medical Record Data}. Based on real hospital records, we constructed the TCMSD-DD-SFT, TCM-WM-Diff-SFT, and TCM-Plan-SFT instruction sets to improve the model's performance on real-world tasks such as syndrome differentiation and disease diagnosis in TCM. The TCMSD-DD-SFT set is designed for TCM syndrome and disease differentiation, aiming to analyze a patient's condition and syndrome type based on their chief complaint, admission details, medical history, and information gathered through TCM's ``Four Diagnoses'' (inspection, listening and smelling, inquiry, and palpation). The TCM-WM-DiffDiag-SFT set includes differential diagnoses in both TCM and western medicine (WM) based on real hospital records. The TCM-Plan-SFT set focuses on providing detailed treatment plans based on a patient's basic condition and diagnosis results. To construct these instruction sets, we manually created a variety of inquiry and response templates, and then employed a weighted random selection approach to generate instruction data from these templates, enhancing the quality and diversity of the instruction sets.

\textbf{Multi-turn Dialogue Data}. To equip the model with conversational capabilities in the medical field, we utilized the DISC-Med-SFT dataset \cite{bao2023disc}. DISC-Med-SFT is a high-quality SFT dataset containing a large number of samples that were reconstructed from existing medical datasets. These data help the model learn medical knowledge, align its behavior with human preferences, and adapt to the distribution of real-world online medical dialogues.

\textbf{Exam Data}. The MLEC-QA dataset is a large-scale Chinese biomedical multiple-choice question-answering dataset \cite{li2021mlec}, sourced from the Chinese Medical Licensing Examination. It covers five subfields: TCM, integrative medicine (combining TCM and Western medicine), clinical medicine, public health, and basic clinical medicine. Based on this dataset, we constructed the MLEC-SFT instruction set to enhance the model's comprehensive question-answering abilities in TCM and healthcare.

\textbf{General Domain Data}. To prevent the loss of the model's general capabilities, we mixed in some general domain data during the SFT process. MS-Agent-Bench\footnote{\url{https://www.modelscope.cn/datasets/iic/MSAgent-Bench}} is a general-purpose multi-domain SFT dataset, from which we randomly selected 65,000 samples. Self-Cognition is a custom instruction set we manually constructed to help the model develop self-awareness capabilities.

The data statistics for the SFT stage are shown in Table \ref{tab:sft-data}. Using this data, we performed supervised fine-tuning on the BianCang TCM base model, ultimately resulting in the BianCang TCM instruction model.

The NLL loss for supervised fine-tuning based on dialogue or question answering can be defined as:
\begin{align}
\mathcal{L}_{\text{SFT}}(\theta) = - \frac{1}{N} \sum_{i=1}^{N} \sum_{t=1}^{T_i} &\log P_\theta(y_t^{(i)} \mid x_1^{(i)}, \\ 
\notag &\ldots, x_{T_i}^{(i)}, y_1^{(i)}, \ldots, y_{t-1}^{(i)}),
\end{align}
where \( x_j^{(i)} \) denotes input tokens (e.g., question or dialogue context) in the \( i \)-th example, whereas \( y_t^{(i)} \) denotes the token at position \( t \) in the response. And, \( P_\theta(\cdot) \) denotes the probability of generating the token at position \( t \) given the input context and previously generated response tokens, computed by the model with parameters \( \theta \) \cite{dai2025language,zhang2025gps}.

\subsection{Model Training Setups}

We train BianCang at two scales, with parameter sizes of 7 billion and 14 billion, based on Qwen2/2.5-7B and Qwen2.5-14B \cite{yang2024qwen2, yang2024qwen25}, respectively. 
We employ full-parameter fine-tuning during both the continual pre-training and supervised fine-tuning stages. We use Scalable lightWeight Infrastructure for Fine-Tuning (SWIFT) \cite{zhao2024swift} as the training framework. 

To ensure training stability, we set the weight decay to 0.1 for the 14B model and 0.01 for the 7B model to prevent overfitting, the max gradient norm to 0.5 to prevent gradient explosion, and the warmup ratio to 0.05 for 14B model and 0.03 for 7B model to maintain smooth training. To balance training costs, we utilize parallel training, gradient checkpointing, fp16 precision, and a gradient accumulation strategy while limiting the model’s maximum response length to 4096 tokens. We reserve 5\% of the training set for validation and use token-level accuracy to validate the model, saving the best checkpoint as the final model. Experiments are conducted on 8 A100 40G GPUs, with both continual pre-training and supervised fine-tuning performed over two epochs. The loss across all training stages successfully converges to a valid range. The continual pretraining process for the 7B model took approximately 174 hours, followed by 119 hours of supervised fine-tuning. In comparison, the 14B model required around 805 hours for continual pretraining and 605 hours for supervised fine-tuning.

\section{Experiments}

The evaluations are conducted from both objective and subjective perspectives. The objective evaluation includes 10 datasets, involving 3 tasks: TCM syndrome differentiation, TCM disease diagnosis, and exams. The subjective evaluation is conducted from 3 dimensions: professionalism, fluency, and safety, using the BC-Analytical dataset constructed by us.

\subsection{Baselines}
\label{baseline_model}
To comprehensively evaluate the performance of our model alongside existing models, we selected a series of LLMs with varying parameters as comparison baselines, including general, medical, and TCM-specific LLMs.

\textbf{GPT-4o} \cite{achiam2023gpt}: GPT-4o is a large-scale, multimodal model developed by OpenAI. It has demonstrated human-level performance across various professional and academic benchmarks. We conduct tests using the API service provided by OpenAI.

\textbf{DeepSeek-V3} \cite{liu2024deepseek}: DeepSeek-V3 is a strong Mixture-of-Experts (MoE) language model with 671B total parameters and 37B activated for each token. It adopts Multi-head Latent Attention (MLA) and DeepSeekMoE architectures, achieving efficient inference and cost-effective training, and is pre-trained on 14.8 trillion diverse and high-quality tokens. DeepSeek-V3 outperforms other open-source models and achieves performance comparable to leading closed-source models.

\textbf{DeepSeek-R1} \cite{guo2025deepseek}: DeepSeek-R1 is the first-generation reasoning model developed by DeepSeek AI, designed to enhance reasoning performance through multi-stage training and cold-start data. DeepSeek-R1 is suitable for tasks that require complex reasoning, such as mathematical problem solving, logical reasoning, and natural language understanding.

\textbf{BianQue2-6B} \cite{chen2023bianque}: BianQue-6B is a ChatGLM-based \cite{du2022glm} LLM finetuned with the self-constructed health conversation dataset BianQueCorpus that consists of multiple turns of questioning and health suggestions polished by ChatGPT. BianQue2-6B expands upon the previous version by incorporating data from pharmaceutical instruction guides, medical encyclopedia knowledge, and ChatGPT distillation instructions, enhancing the model's capabilities in providing recommendations and knowledge retrieval.

\textbf{DoctorGLM-6B} \cite{xiong2023doctorglm}: A large-scale Chinese medical model based on ChatGLM-6B \cite{du2022glm}, fine-tuning on a large amount of medical instruction dataset.

\textbf{BenTsao-Llama-7B} \cite{wang2023huatuo}: A Chinese medical LLM based on Chinese-LLaMA-7B \cite{cui2023efficient}, and fine-tuned on an 8K medical dialogue dataset.

\textbf{ShenNong-TCM-LLM-7B} \cite{zhu2023ShenNong}: A large traditional Chinese medicine model based on Chinese-LLaMA-7B \cite{cui2023efficient}, fine-tuned on the self-constructed ChatMed-TCM dataset.

\textbf{Taiyi-7B} \cite{luo2024taiyi}: A bilingual medical large language model designed for various biomedical tasks, trained on 140 Chinese and English biomedical datasets across 15 BioNLP tasks.

\textbf{WiNGPT2}\footnote{\url{https://github.com/winninghealth/WiNGPT2}}: A large language model developed by Weining Health for the medical domain, aimed at providing intelligent medical Q\&A, diagnostic support, and medical knowledge services to improve the efficiency of diagnosis and the quality of healthcare services. WiNGPT2-7B-Base is based on Qwen-7B \cite{bai2023qwen}, while WiNGPT2-Llama3-8B-Base is based on LLaMA3-8B \cite{dubey2024llama}.

\textbf{HuatuoGPT2} \cite{chen2023huatuogpt}: HuatuoGPT2 employs an innovative domain adaptation method to significantly boost its medical knowledge and dialogue proficiency. We utilized two models, HuatuoGPT2-7B and HuatuoGPT2-13B, which are based on Baichuan2-7B-Base and Baichuan2-13B-Base \cite{yang2023baichuan}, respectively.

\textbf{Sunsimiao-7B} \cite{Sunsimiao}: A Chinese medical large language model based on Qwen2-7B \cite{yang2024qwen2}, fine-tuned on high-quality medical data. It has achieved excellent results in Chinese medical exams.

\textbf{QiZhen-CaMA-13B}\footnote{\url{https://github.com/CMKRG/QiZhenGPT}}: A Chinese medical large language model based on CaMA-13B \cite{knowlm}, and it was trained on the QiZhen medical knowledge base with a Chinese medical instruction set. It performs exceptionally well in Chinese medical scenarios.

\textbf{Zhongjing-13B} \cite{yang2024zhongjing}: A Chinese medical large language model that has undergone end-to-end training, from pre-training to reinforcement learning. In certain dialogue scenarios, it approaches the professional level of expert doctors.  Zhongjing-13B is based on Ziya-LLaMA-13B-v1 \cite{zhang2022fengshenbang}, a general Chinese LLM with 13 billion parameters, trained based on LLaMA.

\textbf{ChiMedGPT-13B} \cite{chimed2024tian}: ChiMed-GPT is a Chinese medical large language model based on Ziya-LLaMA-13B-v2 \cite{zhang2022fengshenbang}. It has undergone comprehensive pre-training, SFT, and RLHF.

\textbf{DISC-MedLLM-13B} \cite{bao2023disc}: A large language model specifically designed for medical and healthcare conversational scenarios. DISC-MedLLM-13B is based on Baichuan-13B-Base\footnote{\url{https://github.com/baichuan-inc/Baichuan-13B}} and trained on the DISC-Med-SFT multi-turn dialogue dataset.

\textbf{Lingdan-13B} \cite{hua2024lingdan}: A traditional Chinese medicine (TCM) large language model based on Baichuan2-13B-Base \cite{yang2023baichuan}, trained on a rich dataset including TCM ancient books, textbooks, and the Chinese Pharmacopoeia.

\textbf{GLM4} \cite{glm2024chatglm}: GLM-4-9B is the open-source version of the latest generation of pre-trained models in the GLM-4 series launched by Zhipu AI. 

\textbf{Qwen2/2.5} \cite{yang2024qwen25}: Qwen2 is a series of open-source multilingual pre-trained and instruction-tuned models based on the Transformer architecture, developed by the Qwen Team. These models have achieved outstanding results across various benchmark evaluations. Qwen2.5 represents a significant upgrade over Qwen2, optimized to meet specific requirements.

\begin{table*}[t]
\caption{Objective evaluation results of traditional Chinese medicine. DI denotes direct inference, CoT denotes chain-of-thought, ZS denotes zero-shot, and FS denotes few-shot. Notably, BianCang outperforms all baseline models with comparable scale (ranging from 6B to 14B), surpassing even GPT-4o and DeepSeek-V3 (671B) on all metrics. Moreover, compared to the reasoning model DeepSeek-R1 (671B), BianCang demonstrates significant advantages across multiple evaluations and achieves comparable results on most metrics.} 
\begin{center}
\renewcommand{\arraystretch}{1.19}
\resizebox{0.95\textwidth}{!}{

\begin{tabular}{l|cccc|cccc|cccc}
 \midrule
\multirow{4}{*}{\bf Model} & \multicolumn{4}{c|}{\bf TCM Syndrome Differentiation} & \multicolumn{4}{c|}{\bf TCM Disease Diagnosis} & \multicolumn{4}{c}{\bf TCM Exam}  \\  \cmidrule(lr){2-5} \cmidrule(lr){6-9} \cmidrule(lr){10-13}
 & \multicolumn{2}{c}{TCMSD} & \multicolumn{2}{c|}{TCMSD-BC} & \multicolumn{2}{c}{TCMDD} & \multicolumn{2}{c|}{TCMDD-BC} & \multicolumn{2}{c}{MLEC-TCM} & \multicolumn{2}{c}{MLEC-CWM} \\ 
 & \multicolumn{2}{c}{Acc.(\%)} & \multicolumn{2}{c|}{Acc.(\%)} & \multicolumn{2}{c}{Acc.(\%)} & \multicolumn{2}{c|}{Acc.(\%)} & \multicolumn{2}{c}{Acc.(\%)} & \multicolumn{2}{c}{Acc.(\%)} \\ 
 & DI& CoT & DI& CoT & DI& CoT & DI& CoT & ZS & FS & ZS & FS \\ \midrule

 GPT-4o~\cite{achiam2023gpt}  & 24.53& 45.21& 16.67&  70.73& 27.83&  54.54& 41.80&  68.33& 74.70 &  76.35& 76.26 &  76.37\\ 
 
   \rowcolor{mygray}DeepSeek-V3~\cite{liu2024deepseek}  & 34.62 & 40.74 & 24.53 & 72.00 & 46.97 & 59.08 & 82.67 & 72.93 & 84.97 & 88.56 & 85.05 & 87.81 \\
   
   DeepSeek-R1~\cite{guo2025deepseek} & 37.17& 55.67 & 25.53 & 76.07 & 50.66 & 80.75 & 79.27 & \bf 94.53 & \bf 92.68 & \bf 93.10 & 90.92 & 90.77 \\
   
 \midrule
 \rowcolor{mygray}BianQue2-6B~\cite{chen2023bianque} & 7.00& 13.87& 15.13&  39.67& 14.76&  21.35& 10.33& 13.93& 4.73 &  3.30& 3.85 &  3.25\\
 DoctorGLM-6B~\cite{xiong2023doctorglm} & 9.17&  9.46& 1.13&  4.87& 11.65&  28.67& 5.60&  17.33& 33.72 &  31.07& 32.86 &  30.09\\
 \rowcolor{mygray}BenTsao-Llama-7B~\cite{wang2023huatuo} & 0.00&  0.11& 0.80&  2.47& 4.50&  9.79& 3.33&  9.33& 23.00 &  21.12& 22.69 &  21.46\\
  ShenNong-TCM-LLM-7B~\cite{zhu2023ShenNong} & 0.18&  1.93& 0.07&  0.67& 7.91&  19.23& 6.20&  14.93& 23.36 &  21.51& 22.58 &  22.65\\
 \rowcolor{mygray}Taiyi-7B~\cite{luo2024taiyi} & 22.57&  22.26& 19.87&  33.20& 23.71&  31.52& 33.47&  54.60& 43.05 &  44.80& 43.93 & 45.16\\
  WiNGPT2-7B-Base & 13.91&  18.19& 15.47&  28.00& 20.87&  33.25& 34.00&  40.73& 54.49 &  55.04& 56.56 &  56.37\\
 \rowcolor{mygray}HuatuoGPT2-7B~\cite{chen2023huatuogpt} & 30.48&  34.09& 38.13&  46.07& 24.61&  39.96& 63.00&  35.60& 47.23 &  50.37& 48.19 &  49.98\\
  Sunsimiao-7B~\cite{Sunsimiao} & 14.53&  18.88& 12.93&  28.87& 27.11&  39.83& 51.60&  49.80& 70.39 &  76.68& 68.45 &  75.40\\
 \rowcolor{mygray}WiNGPT2-Llama3-8B-Base & 4.65&  8.55& 4.67&  17.00& 32.41&  34.98& 49.07&  16.67& 51.12 &  49.69& 52.79 &  52.64\\
  QiZhen-CaMA-13B & 4.39&  6.87& 4.60&  9.20& 10.59&  14.05& 16.13&  14.53& 19.92 &  21.83& 19.33 &  21.50\\
 \rowcolor{mygray}Zhongjing-13B~\cite{yang2024zhongjing} & 0.80&  6.29& 0.80&  16.00& 14.40&  25.65& 13.87&  19.60& 23.03 &  23.00& 23.63 &  22.69\\
 ChiMedGPT-13B~\cite{chimed2024tian} & 1.04&  3.52& 1.87&  5.20& 6.84&  9.42& 5.87&  5.87& 22.45 &  23.23& 23.48 &  26.21\\
 \rowcolor{mygray}DISC-MedLLM-13B~\cite{bao2023disc} & 2.55&  3.54& 0.13&  1.80& 9.86&  16.53& 14.07&  12.13& 37.09 &  35.57& 37.53 &  35.44\\
 HuatuoGPT2-13B & 22.60&  22.06& 23.53&  44.60& 27.73&  41.49& 22.73&  26.67& 49.37 &  50.11&  50.69 & 49.91\\ 
  \rowcolor{mygray}Lingdan-13B~\cite{hua2024lingdan} &  31.68&  23.02& 27.93&  52.00& 42.67&  48.74& 70.93&  59.80&  58.99&  58.18&   57.57& 57.64\\
  \midrule
GLM4-9B~\cite{glm2024chatglm} & 27.42&  30.00& 25.07&   32.53& 36.33&  52.15& 65.20&  74.20& 76.06 &  76.51& 75.93 &  76.00\\
  \rowcolor{mygray}Qwen2-7B~\cite{yang2024qwen2}& 31.74&  27.18& 32.73&  28.40& 41.60&  54.59& 74.87&  77.93& 86.01 &  89.18& 84.45 &  87.89\\
 Qwen2-7B-Instruct& 25.70&  33.41& 14.27&  57.00& 32.87&  52.92& 60.40&  60.13& 83.61 &  84.22& 79.89 &  82.99\\ 
 \rowcolor{mygray}Qwen2.5-7B& 30.44&  21.29& 17.87&  35.73& 23.71&  43.88& 63.87&  71.27&  83.32&  85.52&  82.02&  84.04\\
 Qwen2.5-7B-Instruct& 24.30&  32.19& 9.93&  57.07& 36.29&  51.51& 62.93&  55.53&  78.72&  79.88&  77.27&  78.43\\
 \rowcolor{mygray}  Qwen2.5-14B &  35.62&  25.21&  33.93&   30.13& 24.33&  36.64& 33.33&  32.80&   86.59&  89.93&   87.10&  90.06\\

Qwen2.5-14B-Instruct &  25.94&   35.03& 16.07&   60.00&  38.30&   49.31& 46.27&   53.67&    82.25&   84.81&    81.79&    85.68\\
 \midrule
 \rowcolor{mygray} BianCang-Qwen2-7B (Ours) & 42.14&  30.30&  57.80&   48.00& 43.73&  54.67& 74.73&  80.67&  90.86 &  91.87&  89.08 &  90.36\\
    BianCang-Qwen2-7B-Instruct (Ours) &  68.88&   75.96& 57.33&   75.40&  64.42&   77.71&  \bf 89.07&   85.67&    92.39 &    92.39&   91.14 &   91.48 \\ 

  \rowcolor{mygray}BianCang-Qwen2.5-7B (Ours) &  46.57&  26.72&  52.93&   45.47& 49.80&  53.15& 68.13&  61.73&   86.46&  86.30&   83.93&  85.35\\

    BianCang-Qwen2.5-7B-Instruct (Ours) &  78.90& \bf 82.10& \bf 66.73& \bf 77.73&  73.73&   \bf 82.65&  87.87&   89.40&    90.22&   90.57&    90.32&    90.62\\ 
  \rowcolor{mygray}BianCang-Qwen2.5-14B (Ours) &  43.77&  33.96&  61.93&   53.47& 66.61&  60.39& 82.93&  77.07&   89.28&  90.86&   89.42&  90.58\\

    BianCang-Qwen2.5-14B-Instruct (Ours) &  \bf 79.38&   75.54& 62.27&   70.73&  \bf 77.63&   82.05& 86.33&   88.73&    92.29&   92.29&    \bf 92.75& \bf 92.86\\
  \midrule

\end{tabular}
}
\end{center}

\label{tab:obj-eval-tcm}
\end{table*}

\begin{table*}[htbp]
\caption{Objective evaluation results of Chinese medical exam. Results marked with * are taken from the CMB benchmark leaderboard. ZS denotes zero-shot and FS denotes few-shot. BianCang demonstrates outstanding performance in modern medicine, surpassing all baseline models with comparable scale (ranging from 6B to 14B). Moreover, BianCang achieves results second only to DeepSeek-R1 (671B) in the comprehensive medical benchmark (CMB), highlighting  its robust capabilities in modern medicine.}
\centering
\renewcommand{\arraystretch}{1.19}
\resizebox{0.9\textwidth}{!}{

\begin{tabular}{l|c|cc|cc|cc}
 \midrule
\multirow{3}{*}{\bf Model} &\,\, CMB \,\,& \multicolumn{2}{c|}{\,\,\,\,MLEC-Clinic\,\,\,\,} & \multicolumn{2}{c|}{MLEC-PublicHealth} & \multicolumn{2}{c}{MLEC-Stomatology} \\ 
 & Acc.(\%) & \multicolumn{2}{c|}{Acc.(\%)} & \multicolumn{2}{c|}{Acc.(\%)} & \multicolumn{2}{c}{Acc.(\%)} \\ 
& ZS/*FS  & \,\,\,\,ZS\,\,\,\, & FS & \,\,\,\,\,\,\,\,ZS\,\,\,\,\,\,\,\, & FS & \,\,\,\,\,\,\,\,ZS\,\,\,\,\,\,\,\, & FS \\ \midrule
 GPT-4o   & 59.46* &82.63 & 82.69&81.55 & 82.58& 72.97 &   75.43\\ 
 
 \rowcolor{mygray}DeepSeek-V3~\cite{liu2024deepseek}  & 82.33& 86.83& 89.41 & 85.38 & 87.59 & 79.09 & 81.97 \\
 
   DeepSeek-R1~\cite{guo2025deepseek} & \bf 86.38&92.51 & 92.36&91.42  & 90.40& 87.03 & 86.16 \\
   
 \midrule
 \rowcolor{mygray}BianQue2-6B  & 7.38*& 3.95 &  2.88& 4.26 & 2.86& 4.23 & 2.27\\
 DoctorGLM-6B   & 7.63*&32.08 & 29.79& 34.30 & 30.91& 30.85 & 28.24\\
 \rowcolor{mygray}BenTsao-Llama-7B   & 20.39*&22.60 &  21.80& 22.38 & 20.87& 22.00 & 22.38\\
  ShenNong-TCM-LLM-7B   & 12.19&21.80 & 20.79& 22.38 & 23.09& 22.31 & 21.13\\
 \rowcolor{mygray}Taiyi-7B   & 15.52&56.62 & 55.81& 49.68 & 49.46& 43.86 &44.08\\
  WiNGPT2-7B-Base   & 43.80& 75.85 & 76.87& 68.99 & 65.37& 59.02 & 59.21\\
 \rowcolor{mygray}HuatuoGPT2-7B   & 59.00* & 52.87 & 56.26& 51.46 & 51.08& 45.48 & 48.02\\
  Sunsimiao-7B   & 79.17& 76.39 & 80.94& 72.71 & 76.91& 67.98 & 71.49\\
 \rowcolor{mygray}WiNGPT2-Llama3-8B-Base   & 51.93& 73.18 & 73.00& 66.61 & 66.56& 59.02 & 59.17\\
  QiZhen-CaMA-13B   & 5.07& 19.33 & 21.89& 21.25 & 20.23& 21.89 & 20.68\\
 \rowcolor{mygray}Zhongjing-13B   & 3.44& 25.04 & 23.91& 24.38 & 23.68& 23.21 & 22.12\\
  ChiMedGPT-13B   & 20.68 & 28.49 & 26.85& 26.54 & 24.06& 26.81 & 26.09\\
 \rowcolor{mygray}DISC-MedLLM-13B   & 39.76*& 40.05 & 39.34& 39.64 & 39.21& 38.19 & 35.65\\
  HuatuoGPT2-13B   & 67.85* & 55.31 & 56.26& 55.50 & 55.39& 49.57 &49.38\\ 
  \rowcolor{mygray}Lingdan-13B   & 50.49 &  64.91& 65.77&  58.14& 59.92&  52.10&52.17\\
  \midrule
GLM4-9B   & 66.57& 78.62 & 78.77& 76.43 & 75.89& 65.97 & 65.82\\
  \rowcolor{mygray}Qwen2-7B  & 81.63& 87.63 &  90.63& 82.63 & 86.79& 80.34 & 84.65\\
 Qwen2-7B-Instruct  & 83.45& 85.16 & 83.35& 81.61 & 81.07& 76.29 & 75.88\\ 
  \rowcolor{mygray}Qwen2.5-7B  & 79.60 &  86.65&  88.55&  83.39& 85.17&  78.03& 80.79\\
 Qwen2.5-7B-Instruct  & 79.51 &  82.81& 83.73&  80.96& 80.85&  72.93& 74.40\\
  \rowcolor{mygray}  Qwen2.5-14B & 84.07 &  90.40&  93.13&   86.46& 89.54&  84.31&   88.20\\

Qwen2.5-14B-Instruct & 83.69 &   86.47& 88.02&   83.17&  86.14&   78.94&    82.57\\
 \midrule
 \rowcolor{mygray}BianCang-Qwen2-7B (Ours)   & 83.27&  91.88 & 93.31&  88.57 & 90.72&  85.29 & 88.47\\
    BianCang-Qwen2-7B-Instruct (Ours)   &  84.08&   94.35 &  94.35 &  91.37 &  \bf 91.64&   89.19 &  90.02\\ 

  \rowcolor{mygray}BianCang-Qwen2.5-7B (Ours) & 80.13 &  90.43&  91.32&   85.65& 87.22&  82.19&   82.65\\

    BianCang-Qwen2.5-7B-Instruct (Ours) & 80.71  &   93.40& 93.43&   89.91&  89.91&   86.43&   86.77\\ 
  \rowcolor{mygray}BianCang-Qwen2.5-14B (Ours) & 84.34 &  91.70&  93.37&   87.92& 89.97&  86.16&   87.94\\ 

    BianCang-Qwen2.5-14B-Instruct (Ours) & 83.80 &   \bf 94.74& \bf 94.97&   \bf 91.86&  91.53&   \bf 90.43&     \bf 90.51\\
  \midrule

\end{tabular}

}
\label{tab:obj-eval-exam}
\end{table*}

\subsection{Objective Evaluation Details and Datasets}
\label{obj_eval_detail}
The objective evaluation primarily tests the model's capabilities in three areas: TCM syndrome differentiation, TCM disease diagnosis, and mastery of TCM and medical knowledge.

For the TCM syndrome differentiation task, we used the TCMSD and TCMSD-BC datasets. The TCMSD dataset was adapted from the TCM-SD dataset \cite{mucheng2022tcm}, where we constructed queries and responses using fields from the test set of TCM-SD dataset. The TCMSD-BC dataset was built based on proprietary medical records. The TCM syndrome differentiation task is divided into two test modes: direct inference (DI) and chain of thought (CoT), with accuracy serving as the evaluation metric. 

For the TCM disease diagnosis task, we used the TCMDD and TCMDD-BC datasets. Similarly, the TCMDD dataset was adapted from the test set of the TCM-SD dataset \cite{mucheng2022tcm}, while the TCMDD-BC dataset was constructed based on proprietary medical records. Like the syndrome differentiation task, the disease diagnosis task is evaluated using two test modes: DI and CoT, with accuracy as the evaluation metric.
It should be noted that the inputs for TCM syndrome differentiation and disease diagnosis tasks tend to be lengthy. However, certain models (e.g., DoctorGLM and BianQue) are constrained by a limited maximum context length. To ensure fairness, both the DI and CoT evaluation modes for these tasks employ a zero-shot approach to avoid context overflow.

To assess the model's mastery of TCM and medical knowledge, we applied the MLEC-QA \cite{li2021mlec} and Chinese Medical Benchmark (CMB\footnote{\url{https://cmedbenchmark.llmzoo.com/static/leaderboard.html}}) \cite{wang2024cmb} datasets. The MLEC-QA dataset is divided into five subsets: traditional Chinese medicine, traditional Chinese medicine combined with western medicine, clinic, public health, and stomatology. The CMB dataset includes six categories: physician exam, nursing exam, pharmacist exam, medical technician exam, professional knowledge exam, and medical postgraduate entrance exam. The average score across these six categories is used as the final result. The examination tasks are evaluated in two test modes: zero-shot and few-shot, with accuracy as the evaluation metric.

We evaluated the model using the exact match (EM) method. Specifically, we first extracted the answer segment from the model's response and then performed a word-by-word comparison between the extracted answer and the reference answer. A match was considered successful only if the two answers were entirely identical. Finally, we computed the proportion of successful matches to derive the accuracy, which served as the metric for evaluating the model's performance.

\subsection{Objective Evaluation Results}
The objective evaluation results in the field of Traditional Chinese Medicine (TCM) are presented in Table \ref{tab:obj-eval-tcm}. The experimental results demonstrate that BianCang achieves strong overall performance across syndrome differentiation, disease diagnosis, and TCM examination tasks. Most existing baselines exhibit weak analytical capabilities in syndrome differentiation and disease diagnosis. We identify three key factors contributing to this limitation: (1) their foundational models are often relatively weak, leading to insufficient general reasoning capacity; (2) TCM tasks require systematic knowledge and diagnostic experience that most models lack due to limited exposure to real-world TCM data; and (3) although some models utilize continual pre-training and instruction tuning, they often fail to effectively align these stages, resulting in inconsistencies between learned knowledge and fine-tuning objectives.

BianCang demonstrates robust analytical performance across both standard and custom datasets. For example, BianCang-Qwen2.5-7B-Instruct achieves an accuracy of 78.90\% on the TCMSD dataset under direct inference (DI), outperforming the foundational model Qwen2.5-7B by over 48 percentage points, and surpassing the strongest TCM baseline Lingdan-13B by a similar margin. Under the chain-of-thought (CoT) setting, its accuracy rises to 82.10\%, showing further improvements over both base and SOTA baselines. Comparable trends are observed on the TCMSD-BC, TCMDD, and TCMDD-BC datasets, confirming that BianCang generalizes well across different TCM task formats. Moreover, in the TCM examination tasks, BianCang achieves performance close to DeepSeek-R1 (670B), and notably achieves the highest accuracy in the Integrated Chinese and Western Medicine (MLEC-CWM) subtask, reflecting its comprehensive and structured grasp of both traditional and modern diagnostic paradigms.

It is worth noting that although BianCang achieves competitive results across the board, it is marginally outperformed by DeepSeek-R1 in three specific cases: TCMDD-BC (BianCang: 89.40\%, DeepSeek-R1: 94.53\%), MLEC-TCM (zero-shot) (92.39\% vs. 92.68\%), and MLEC-TCM (few-shot) (92.39\% vs. 93.10\%). These differences may be attributed to DeepSeek-R1's massive parameter scale (670B), which allows it to better capture implicit reasoning patterns and general medical knowledge, particularly in few-shot and CoT-driven tasks. Nevertheless, BianCang attains highly competitive or superior performance with a much smaller 7B/14B model, underscoring the efficacy of our domain-specific training approach. Interestingly, we also observe that BianCang-Qwen2.5-7B-Instruct outperforms the larger 14B-Instruct variant in certain tasks, such as TCMSD (CoT: 82.10\% vs. 75.54\%) and TCMDD (CoT: 82.65\% vs. 82.05\%). This result may be due to larger models overinterpreting instructions, leading to unnecessary reasoning that reduces diagnostic accuracy. Smaller models tend to follow instructions more directly and focus on key information. Moreover, large models are more prone to training instability and forgetting during fine-tuning. Finally, we observe that the performance gap between DI and CoT inference in BianCang is considerably smaller than in other models—for instance, DeepSeek-R1 exhibits a 50.5-point gap on TCMSD-BC (DI: 25.53\%, CoT: 76.07\%), while BianCang-Qwen2.5-7B-Instruct exhibits a smaller gap of only 11 points (DI: 66.73\%, CoT: 77.73\%). This suggests that BianCang has internalized key diagnostic reasoning patterns through domain-specific training and does not rely heavily on external CoT prompting. In contrast, general-purpose models such as DeepSeek appear to depend more on explicit CoT reasoning to compensate for a lack of domain alignment, leading to significantly weaker performance in direct inference.

Additionally, we report more evaluation results on the Chinese medical examination in Table \ref{tab:obj-eval-exam}. As illustrated in Table \ref{tab:obj-eval-exam}, BianCang not only performs well in the TCM domain but also demonstrates outstanding performance in modern medicine, surpassing all baselines with comparable scale (ranging from 6B to 14B) in Clinic, Public Health, and Stomatology examinations. Moreover, BianCang achieves results second only to DeepSeek-R1 (671B) in the comprehensive medical benchmark (CMB), underscoring its robust capabilities across both TCM and modern medicine. 

\begin{figure*}[htbp]
\centering
\includegraphics[width=1\textwidth]{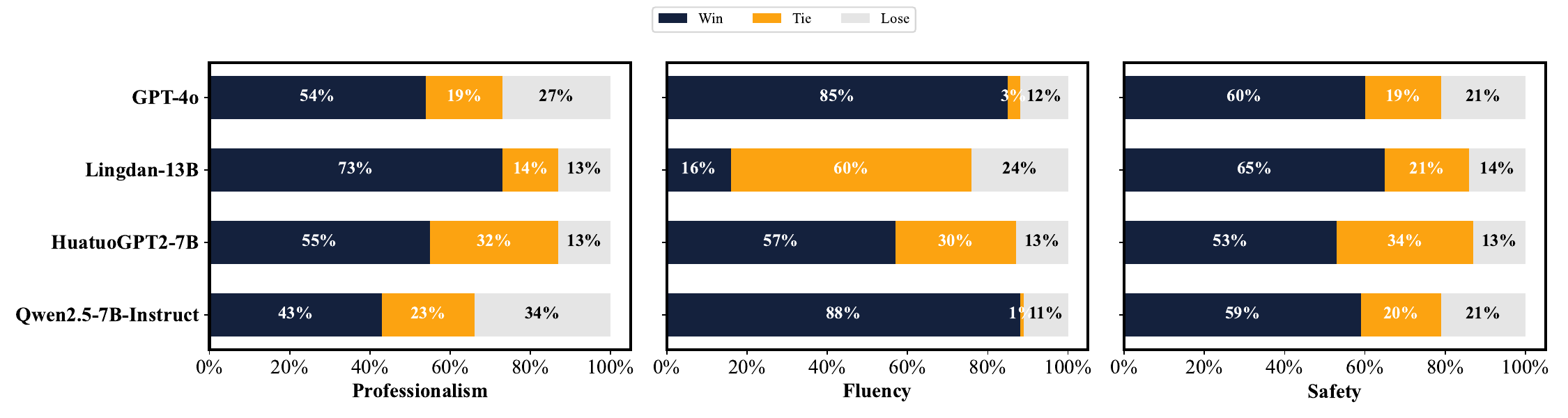}
\caption{Subjective evaluation results of BianCang-Qwen2.5-7B-Instruct and other baseline models in terms of professionalism, fluency, and safety. The test dataset used is BC-Analytical.}
\label{fig:subjective}
\end{figure*}

\begin{table*}[htbp]
\caption{Performance comparison of one-stage and two-stage training methods. The one-stage approach involves supervised fine-tuning of the vanilla model only. The two-stage approach first performs continual pretraining on the vanilla model, followed by supervised fine-tuning. The training data and parameter settings for supervised fine-tuning are identical.}
\begin{center}
\resizebox{0.88\textwidth}{!}{
 \begin{tabular}{c|cccc|cccc|ccc}
\hline
\multirow{4}{*}{\bf Model} & \multicolumn{4}{c|}{\bf TCM Syndrome Differentiation} & \multicolumn{4}{c|}{\bf TCM Disease Diagnosis} & \multicolumn{3}{c}{\bf Exam}  \\  
\cline{2-12}
 & \multicolumn{2}{c}{TCMSD} & \multicolumn{2}{c|}{TCMSD-BC} & \multicolumn{2}{c}{TCMDD} & \multicolumn{2}{c|}{TCMDD-BC} & \multicolumn{2}{c}{MLEC-Avg} & CMB \\ 
 & \multicolumn{2}{c}{Acc.(\%)} & \multicolumn{2}{c|}{Acc.(\%)} & \multicolumn{2}{c}{Acc.(\%)} & \multicolumn{2}{c|}{Acc.(\%)} & \multicolumn{2}{c}{Acc.(\%)} & Acc.(\%) \\ 
 & DI & CoT & DI & CoT & DI & CoT & DI & CoT & ZS & FS & ZS  \\ 
 \hline
 Vanilla  & 30.44 & 21.29 & 17.87 & 35.73 & 23.71 & 43.88 & 63.87 & 71.27 & 82.68 & 84.81 &79.60 \\ \hline
  One Stage & 77.96 & 65.13 & 60.27& 64.60 & 73.00 & 77.14 & 85.73 & \bf 89.73 & 90.04 & 87.26 & 79.90\\ \hline
 Two Stage  & \bf 78.90 & \bf 82.10 & \bf 66.73 & \bf 77.73 & \bf 73.73 & \bf 82.65 & \bf 87.87 & 89.40 & \bf 90.06 & \bf 90.26 & \bf 80.71 \\ 
 \hline
\end{tabular}
}
\end{center}

\label{tab:ablation}
\end{table*}

\subsection{Subjective Evaluation Details and Datasets}
\label{sbj_eval_detail}
We used the BC-Analytical dataset for the subjective evaluation of the model. The BC-Analytical dataset, which we constructed using proprietary medical records, includes 50 complex TCM cases. Each case is accompanied by three questions: disease diagnosis and syndrome differentiation analysis, differential diagnosis, and treatment plan formulation. We evaluate the model from three dimensions: professionalism, fluency, and safety, using win rate, tie rate, and loss rate as the evaluation metrics. All evaluations are conducted by two hospital physicians specializing in TCM cardiology. Following prior research \cite{yang2024zhongjing, jia2025qibo}, we redefined the evaluation criteria for professionalism, fluency and safety. The specific evaluation standards are as follows:

The evaluation criteria for fluency and safety are consistent with previous research \cite{yang2024zhongjing, jia2025qibo}, and the professionalism evaluation criteria are as follows: (1) Must have an accurate understanding of the problems and needs from doctors or patients in order to provide relevant responses and advice; (2) must correctly predict the patient's disease or syndrome and provide supporting rationale when providing diagnostic support; (3) must correctly interpret the medical knowledge involved.

\subsection{Subjective Evaluation Results}
For the subjective evaluation, as shown in Fig. ~\ref{fig:subjective}, we selected four models with strong overall performance in objective evaluation as baselines to reduce human evaluation costs. 

The results indicate that our model demonstrates a higher level of professionalism in analyzing real TCM cases, which is largely due to its well-developed TCM knowledge system and authentic diagnostic experience. In terms of fluency, our model can generate more authentic, TCM-styled outputs, surpassing most baseline models. In terms of safety, such as in diagnostic or medication recommendation scenarios, our model consistently outperformed the baseline models. Overall, our model's responses received greater approval from TCM experts.

\subsection{Effectiveness of Two-stage Training Strategy}
We conducted a set of comparative experiments to illustrate the necessity and effectiveness of the two-stage training strategy. The experimental results are shown in Table \ref{tab:ablation}. ``Vanilla" refers to the foundational model, Qwen2.5-7B. ``One Stage'' represents directly applying supervised fine-tuning (SFT) to the model using an instruction set tailored for TCM downstream tasks. ``Two Stage'' represents first injecting knowledge through continual pre-training and then applying supervised fine-tuning on downstream tasks to maximize consistency between the instructions and the model's internal knowledge during fine-tuning. The SFT data and training parameter settings used in both One Stage and Two Stage are consistent.

The experimental results indicate that our two-stage training strategy is highly effective when the foundational model itself lacks sufficient knowledge. Taking the TCM syndrome differentiation task as an example, we observe that the foundational model performs poorly on this task, highlighting its lack of TCM-specific knowledge in syndrome differentiation. In such cases, directly applying SFT to the foundational model may result in limited performance gains due to inconsistencies between the instructions and the model's internal parameter knowledge. In contrast, with the two-stage training strategy, the consistency between the instructions and the model's internal parameter knowledge is enhanced, leading to a substantial improvement in the model's performance on the TCM syndrome differentiation task. This improvement is especially significant in chain-of-thought scenarios, where effective reasoning depends on the model’s foundational knowledge.

\section{Conclusion}
In this work, we introduce BianCang, a large language model specifically designed for traditional Chinese medicine (TCM), to address challenges unique to TCM large language models. We constructed a comprehensive pre-training corpus and a diverse set of supervised fine-tuning instructions, and employed a two-stage training strategy to train BianCang, which first injects domain-specific knowledge and then aligns it through targeted stimulation.
Experimental results robustly demonstrate the effectiveness and superiority of BianCang.

\section*{Acknowledgment}
This work was supported by the National Natural Science Foundation of China (No.62376130 and No.62402258), Taishan Scholars Program (No.TSPD20240814 and No.TSQN202507242), Shandong Provincial Natural Science Foundation (No.ZR2022MF243), Program of New Twenty Policies for Universities of Jinan (No.202333008), Pilot Project for Integrated Innovation of Science, Education, and Industry of Qilu University of Technology (Shandong Academy of Sciences) (No.2025ZDZX01).

\section*{Clinical Relevance and Impact}
The clinical relevance and impact of BianCang can be discussed from several key perspectives. First, BianCang significantly improves the accuracy of traditional Chinese medicine (TCM) syndrome differentiation and disease diagnosis in real-world scenarios, demonstrating its potential as a clinical diagnostic aid. When combined with retrieval-augmented generation (RAG) technology, BianCang provides TCM practitioners with reliable and accurate prescriptions, thus reducing the cognitive load associated with memorization.

Secondly, BianCang offers potential for enhancing the training and educational resources for TCM students and practitioners. With its superior performance in tasks involving real clinical scenarios and examinations, the model provides a reliable and accessible reference tool that can enrich the learning process.

Moreover, BianCang's effectiveness in handling both pure TCM tasks and integrative medicine tasks (combining TCM with Western medicine) expands its applicability, demonstrating significant utility in contemporary healthcare environments where integrative approaches are increasingly valued.

Lastly, the open-source nature of BianCang ensures broad accessibility, enabling widespread clinical use, further validation, and continuous improvement by the global community. This democratization of specialized TCM knowledge potentially accelerates research, fosters collaboration, and elevates the overall standard of patient care in TCM practice.

\section*{Ethical Considerations}

In the development and training of BianCang, rigorous measures were implemented to ensure the ethical use of patient data. All electronic medical records (EMRs) utilized in this study were derived from real-world patient data; however, comprehensive data anonymization and desensitization techniques were applied to safeguard patient privacy. Specifically, all personally identifiable information (PII), including names, addresses, contact details, and other sensitive data, was either removed or obfuscated. The training dataset was strictly anonymized, ensuring no information could be used to identify individual patients. This approach guarantees full compliance with ethical guidelines regarding patient privacy and confidentiality.
\\ 
\indent Additionally, the resources and data associated with this project are solely intended for academic research purposes. BianCang is designed as a language model-based intelligent assistant aimed at supporting research and providing insights into Traditional Chinese Medicine (TCM)-related topics. However, it is crucial to acknowledge the model's inherent limitations, as it cannot guarantee the accuracy of all responses. BianCang is not a substitute for professional medical diagnosis or advice from licensed TCM or Western medicine practitioners. Users are strongly advised against relying exclusively on the model's outputs for medical decision-making.
\\
\indent Given the potential serious consequences of inaccuracies in medical data, we emphasize the importance of exercising caution when interpreting and utilizing the information generated by the model. Users should always consult qualified healthcare professionals or seek medical attention at a hospital when necessary. BianCang is intended to complement, rather than replace, the expertise of medical professionals.\\
\indent Furthermore, considering the potential safety risks in the application of TCM, we emphasize the following precautions: (1) Herbal compatibility safety, such as traditional contraindications including Shiba Fan (the "Eighteen Incompatibilities") and Shijiu Wei (the "Nineteen Antagonisms"), should be carefully considered; (2) Medication safety during pregnancy, as certain herbs may pose risks to both the mother and fetus; (3) Diet-drug interactions, which are important in TCM practice and require attention to dietary restrictions and contraindications; and (4) Professional supervision, where all medication use should be guided by qualified TCM or medical professionals. Outputs from the model are intended for research purposes only and must not be used as a basis for actual medical treatment, to avoid potential serious consequences.

\section*{References}
\bibliographystyle{ieeetr}
\bibliography{main}
\end{CJK}
\end{document}